\documentclass[conference]{IEEEtran}
\usepackage{blindtext, graphicx}
\usepackage[utf8]{vietnam}
\usepackage{multirow}
\usepackage[table,xcdraw]{xcolor}
\ifCLASSINFOpdf
\else
\fi
\begin{document}

\title{Dialogue Act Segmentation for Vietnamese Human-Human Conversational Texts}
\author{\IEEEauthorblockN{Thi--Lan Ngo}
\IEEEauthorblockA{University of Information and Communication Technology\\
Thainguyen University (TNU)\\
Email: ntlan@ictu.edu.vn}
\IEEEauthorblockN{Khac--Linh Pham}
\IEEEauthorblockA{University of Engineering and Technology\\
Vietnam National University, Hanoi (VNU)\\
Email: phamkhaclinh2017@gmail.com}
\IEEEauthorblockN{Minh--Son Cao}
\IEEEauthorblockA{University of Engineering and Technology\\
Vietnam National University, Hanoi (VNU)\\
Email: soncm\_58@vnu.edu.vn}
\and
\IEEEauthorblockN{Son--Bao Pham}
\IEEEauthorblockA{University of Engineering and Technology\\
Vietnam National University, Hanoi (VNU)\\
Email: sonpb@vnu.edu.vn}
\IEEEauthorblockN{Xuan--Hieu Phan}
\IEEEauthorblockA{University of Engineering and Technology\\
Vietnam National University, Hanoi (VNU)\\
Email: hieupx@vnu.edu.vn}
}


\maketitle

\renewcommand{\abstractname}{Abstract}
\begin{abstract}

Dialog act identification plays an important role in understanding conversations. It has been widely applied in many fields such as dialogue systems, automatic machine translation, automatic speech recognition, and especially useful in systems with human-computer natural language dialogue interfaces such as virtual assistants and chatbots. The first step of identifying dialog act is identifying the boundary of the dialog act in utterances. In this paper, we focus on segmenting the utterance according to the dialog act boundaries, i.e. functional segments identification, for Vietnamese utterances. We investigate carefully functional segment identification in two approaches: (1) machine learning approach using maximum entropy (ME) and conditional random fields (CRFs); (2) deep learning approach using bidirectional Long Short-Term Memory (LSTM) with a CRF layer (Bi-LSTM-CRF) on two different conversational datasets: (1) Facebook messages (Message data); (2) transcription from phone conversations (Phone data). To the best of our knowledge, this is the first work that applies deep learning based approach to dialog act segmentation. As the results show, deep learning approach performs appreciably better as to compare with traditional machine learning approaches. Moreover, it is also the first study that tackles dialog act and functional segment identification for Vietnamese.

\end{abstract}


\begin{IEEEkeywords}
Dialog act segmentation, functional segment, Vietnamese conversation.
\end{IEEEkeywords}

\IEEEpeerreviewmaketitle
\section{Introduction}
\label{sec:intro}
Automatic recognition of user intent from utterances in their interaction with systems through the conversational interface is a very challenging task that has attracted a lot of attention from research community for two decades. The goal is to design methods to make computers interact more naturally with human beings. Identifying dialog acts (DAs) within an utterance, i.e. identifying its illocutionary act of communication, plays a key role in understanding user's intent. Because, \textit{``Dialog act is a communicative activity of dialog participant, interpreted as having a certain communicative function and semantic content"} \cite{Bunt:2012}. It presents meaning of utterances at the discourse level. It is a complementary process to concept extraction. Therefore, it is essential for the complete understanding of conversations. It is important for many applications: dialogue systems, automatic translation machine \cite{Tanaka:1999}, automatic speech recognition, etc \cite{wang:2005} \cite{Kral:2012} and has been studied in various languages such as English, Chinese, Arabic, Czech, Korean. Whilst in Vietnamese languages, dialog act has only been studied in linguistics, our work in this paper is a preliminary study about automatic identification of dialog act, as well as dialog act segmentation.

Prior to DA identification, utterances must be segmented according to DA boundaries. In the past, there have been studies of DA segmentation such as Umit Guz et al. implemented DA segmentation of speech using multi-view semi-supervised learning \cite{Guz:2010}; Jeremy Ang et al. explored DA segmentation using simple lexical and prosodic knowledge sources \cite{Jeremy:2005}; Warnke et al. calculated hypotheses for the probabilities exceeded a predefined threshold level in VERBMOBIL corpus \cite{Warnke:1997}; Silvia Quarteroni et al. segmented human-human dialog into turns and intra-turn segmentation into DA boundaries using CRFs to learn models for simultaneous segmentation of DAs from whole human-human spoken dialogs \cite{Quarteroni:2011}. These studies segmented turns into  sentence unit to do dialog act segmentation. In my work, different from those studies, we segment utterances into the smallest meaningful units -- \textit{``functional segment"} unit. According to ISO 24617-2 standard about Dialog Act, a functional segment (FS) is defined as ``minimal stretch of communicative behavior that have a communicative function"  \cite{Bunt:2012}. For example, in the utterance ``xin chào cậu khỏe chứ" (``hello are you fine"), there are two functional segments: ``xin chào" (``hello”) (its dialog act is greeting), and ``cậu khoẻ chứ" (``are you fine”) (its dialog act is check question). We investigate thoroughly functional segment identification in two approaches: (1) machine learning approach with ME, CRF; (2) deep learning approach with Bi--LSTM--CRF. Recently, ME, CRF and Bi–LSTM–CRF have been applied to a variety of sequence labeling and segmentation tasks in Natural Language Processing and have achieved state-of-the-art results \cite{Huang:2015}. Therefore, we expect that these methods apply to the FS identification task for Vietnamese can make similar successes. To do the task, we first build two annotated corpus from Facebook messages and transcription from phone conversations. For a careful evaluation, different ME, CRF and Bi--LSTM--CRF models were trained and their results are compared and shown contrast with each other. Moreover, we also show the characteristics of two different conversational data sets and their effect on the experimental results of the task of the dialog act segmentation task.

We can summary our main contributions in this paper in two aspects:
\begin{itemize}
    \item First, we built two Vietnamese conversational text datasets which are segmented into FSs based on FS concept from the ISO standard and ready to contribute to the DialogBank \footnote{https://dialogbank.uvt.nl/} for Vietnamese.  We also built online chat dictionary which contains abbreviations, slang words and teen code and Vietnamese local dialect dictionary.
    \item Second, two machine learning techniques and a deep learning technique are applied and compared on the task of automatic dialog act segmentation. Deep learning technique is also applied for the first time to dialog act segmentation. The results of the deep learning technique are very promising, opening up a new way to approach dialog act segmentation and dialog act in general for applications for future studies.
\end{itemize}

The rest of the paper is organized as follows: Section \ref{sec:background} presents briefly background about FS formation in Vietnamese conversational texts and units of a dialogue. In Section \ref{sec:corpus} we describe our two human-human conversation corpus. We also discuss the impact of our conversational data sets to the functional segment identification task in this section. We describe quickly the two learning models ME, CRF and the deep learning model, Bi--LSTM--CRF for labeling and segmenting FS in Section \ref{sec:method}. Section \ref{sec:evaluation} mainly presents the framework of using MEs, CRFs, Bi--LSTM--CRF for Vietnamese FS segmentation and result comparison and evaluation. Finally, Section \ref{Sec:concl} shows some conclusions and the work that need research in the future.

\section{Backgroud: Functional segment and units of a dialogue}
\label{sec:background}
DAs are extended from the speech act theory of Austin \cite{Austin:1975} and Searle \cite{Searle:1975} to model the conversational functions that utterances can perform. It is the meaning of an utterance at the level of illocutionary force, such as statement, question and greeting. Detection of dialog acts need to perform: 1) the segmentation of human--human dialogues into turns, 2) the intra-turn segmentation into DA boundaries, i.e. functional segment identification and 3) the classification of each segment according to a DA tag \cite{Ramacandran:2013}.


In which, ``turn", ``dialog act", ``functional segment" terms are defined slightly different between  different domains and different purposes. But these are standardized and united in ISO standards as follows:

\subsection*{Turn:} A ``turn" is definite as \textit{``stretch of communicative activity produced by one participant who occupies the speaker role bounded by periods where another participant occupies the speaker role"}. Dialogue participants (sender, addressee) normally take turns in conversation. Several utterances from one of the dialogues in our corpus are shown as examples of  \textit{Turn}, \textit{Message}, and \textit{Functional segment} in Table \ref{Tab:Message} and Table \ref{Tab:phone}. \\
In our Message data, a turn is seen as a collection of continuous messages sent by one participant. In which, a message is defined as a group of words that are sent from one dialogue participant to the other. For instance, turn $t_2$ includes four messages $ms_2$, $ms_3$, $ms_4$, $ms_5$ (Table \ref{Tab:Message}).   

\subsection*{Functional segment:}
A functional segment is the \textit{``minimal stretch of communicative behavior that has a communicative function”}, \textit{``minimal in the sense of not including material that does not contribute to the expression of the function or the semantic content of the dialogue act"} \cite{Bunt:2012}.  A functional segment may be shorter than turns and continuous, for example as in Table \ref{Tab:Message}, $t_1$ includes two functional segments ${fs}_1$ and  ${fs}_2$. A functional segment may be discontinuous, with examples such as ${fs}_4$ and ${fs}_{10}$. ${fs}_5$ is nested within ${fs}_4$. 
In addition, functional segment ${fs}_{10}$ is combined from two messages, ${fs}_8$ overlaps ${fs}_{10}$. Thus, we can see that a functional segment may be continuous, may be discontinuous, may be overlapped and nested. The detailed explanation of the types of FS is presented in \cite{Bunt:2011} and the ISO 24617-2 standard.
\subsection*{Dialog Act:}
DA is \textit{``communicative activity of a dialogue participant, interpreted as having a certain communicative function and semantic content"}. For example: 
\begin{center} 
``xin chào cậu khoẻ chứ" (``hello are you fine”)
\end{center}
DAs of ``xin chào" (hello) are \textit{Greeting} and \textit{Opening}. DA of ``cậu khoẻ chứ" (``are you fine”) is \textit{Check Question}.
\renewcommand\tablename{Table}
\begin{table*}[ht]
\renewcommand{\arraystretch}{1.3}
\centering
\caption{Examples of functional segment and turn in Message data.}
\label{Tab:Message}
\begin{tabular}{p{1.2cm}|p{4.2cm}|p{4cm}|p{4.2cm}|p{1.4cm}}
\hline
\textbf{Participants} & \textbf{Messages}   & \textbf{Turns}  & \textbf{Functional segments} & \textbf{Type}\\ \hline
\multirow{2}{*}{S} & \multirow{2}{*}{}{Đây là đề tài chung tôi sẽ hưỡng dẫn bạn dần dần :) (This is the general topic I will guide you gradually :) ) (ms1)} & \multirow{2}{*}{}{Đây là đề tài chung tôi sẽ hướng dẫn bạn dần dần :) (This is the general topic I will guide you gradually :) )  (t1)}   & Đây là đề tài chung (This is the general topic) (fs1) & continuous \\ \cline{4-5} 
& & & Tôi sẽ hướng dẫn bạn dần dần :) (I will guide you gradually :)) (fs2)  & continuous \\ \hline
A & uhhhhhh nhưng thời gian (Yessssss, but the time) (ms2) & \multirow{4}{*}{}{uhhhhhh nhưng thời gian hic hic ngắn quá sợ k làm đc (Yessssss, but the time is too short I am afraid of can not done) (t2)} & uhhhhhh (fs3) & continuous \\ \cline{1-2} \cline{4-5} 
A & hic hic (ms3) & & nhưng thời gian ngắn quá (but the time is too short) (fs4) & discontinuous \\ \cline{1-2} \cline{4-5} 
A & ngắn quá (too short) (ms4)  &   & hic hic (fs5) & nested \\ \cline{1-2} \cline{4-5} 
A & sợ k làm đc (I am afraid of can not done) (ms5) & & sợ k làm đc (I am afraid of can not done) (fs6) & continuous\\ \hline
S   & Cậu còn chưa bắt đầu mà đã sợ rồi (You have not started yet, have you been afraid) (ms6) & Cậu còn chưa bắt đầu mà đã sợ rồi (You have not started yet have been afraid) (t3) & Cậu còn chưa bắt đầu mà đã sợ rồi (You have not started yet have been afraid) (fs7) & continuous                   \\ \hline
A & chưa bắt đầu hic :3 cái gì? (not started yet hic :3 what? )(ms7) & \multirow{4}{*}{}{chưa bắt đầu hic :3 cái gì ? tôi đang làm rồi mà (not started yet hic :3 what? I am doing ) (t4)}  & chưa bắt đầu (fs8) & overlap \\ \cline{1-2} \cline{4-5} 
\multirow{3}{*}{A} & \multirow{3}{*}{Tôi đg làm rồi mà :) (ms8)} &  & hic :3 (fs9)  & continuous \\ \cline{4-5} 
&& & Chưa bắt đầu cái gì? (not started yet hic :3 what?) (fs10)  & overlap and  discontinuous \\ \cline{4-5} 
& & & Tôi đg làm rồi mà :) ( I am doing) (fs11)  & continuous  \\ \hline
\end{tabular}
\end{table*}
\renewcommand\tablename{Table}
\begin{table*}[ht]
\centering
\caption{Examples of functional segment and turn in Phone data.}
\label{Tab:phone}
\begin{tabular}{p{1.3cm}|p{7cm}|p{6cm}|p{1.4cm}}
\hline
\textbf{Participants} & \textbf{Turn} & \textbf{Functional segment}   & \textbf{Type}   \\ \hline
\multirow{2}{*}{S}  & \multirow{2}{*}{}{ở \textless no speech\textgreater \ ở quê có những đặc sản gì vậy anh (in  \textless no speech\textgreater \ in home town What is the specialty) (t1) }   & ở (in) (fs1) \    & continuous  \\
  &  & ở quê có những đặc sản gì vậy anh  (in home town What is the specialty) (fs2) & continuous  \\ \hline
\multirow{3}{*}{A} & \multirow{3}{*}{}{cái này a - ủa cái này thì củng nói thật chứ nhiều đặc sản lắm  \textless no speech\textgreater \  đặc sản quê hương là mỗi nơi mỗi khác  \textless no speech\textgreater \ (About this  I – oh about this then being honest there are a lot specialties specialties of each country is different) (t2) }   & cái này a - (this) (fs3)  & continuous    \\
  &  & ủa cái này thì củng nói thật chứ nhiều đặc sản lắm  (About this  I – oh about this then being honest there are a lot specialties) (fs4)& continuous  \\
  &  & đặc sản quê hương là mỗi nơi mỗi khác ( specialties of each country is different) (fs5)& continuous   \\ \hline
\multirow{2}{*}{S} & \multirow{2}{*}{dạ vâng ạ (yes yes) (t3)\textless no speech\textgreater \    } & dạ (yes) (fs6) & continuous   \\ 
&  & vâng ạ (yes) (fs7) & continuous   \\ \hline
\multirow{6}{*}{A}    & \multirow{6}{*}{}{ở trên này thì có là là nói chung là như là ở sông thì có cá sông nả  \textless no speech\textgreater \  cá sông là tuyệt vời nhức hes chơ mà dưới biển thì có cá biển \textless laugh\textgreater \  nhưng mà ở sông thì lại lại lại chuộng cấy cá (Over here there are are in general  there are are like river has river fishes  \textless nospeech\textgreater river fishes are the best but in sea we also have sea fishes but near river also also also prefer sea fishes  \textless laugh\textgreater ) (t4) } & ở trên này thì có như là ở sông thì có cá sông nả    (Over here there are are in general  there are are like river has river fishes ) (fs8)          & \multicolumn{1}{l}{discontinuous} \\
 &  & là là nói chung là    (is is in general) (fs9)    & \multicolumn{1}{l}{nested}  \\
 &  & cá sông là tuyệt vời nhức (river fishes are the best) (fs10)                & \multicolumn{1}{l}{continuous}    \\
 &  & chơ mà dưới biển thì có cá biển  (but in sea we also have sea fishes )(fs11)        & \multicolumn{1}{l}{continuous}    \\
 &  & nhưng mà ở sông thì lại chuộng cấy cá (but near river also also also prefer sea fishes ) (fs12)    & \multicolumn{1}{l}{continuous}   \\
 &  & lại lại           (fs13)                        & \multicolumn{1}{l}{nested}   \\ \hline
\end{tabular}
\end{table*}




\section{Corpus Building: Message data \& Phone data}
\label{sec:corpus}
In Vietnamese, there is no publicly available standard corpus. Therefore we need to build first a reference corpus for training and evaluation. For this work, we have to build two corpora of data from human-human conversations in various domains. One is chat texts and other is spoken texts. 
\subsection{Message corpus}
Our Message data set is collected from Facebook messages of 20 volunteers. The data set contains 280 human-human Vietnamese dialogues in any topics with a total number of 4583 messages. The average length of dialogues is 16.4 messages. The data set was independently labeled by three annotators. The agreement score of our data set achieved 0.87 Fleiss’ kappa measure \cite{fleiss:1971}. As observed from our data, there are some challenges as follows: 


\begin{enumerate}
\item The data is very noisy because it contains many acronyms, misspellings, slang, and emoticons. These informal natures of chat text, which make conventional features such as punctuation mark, part--of--speech (POS), syntax of sentence and capitalization, are not reliable. Text message conversations are often written with non-standard word spellings. While some of them are unintentional misspellings, many of them are purposely produced, for example,\\
\indent S:  ``đi chơi điiiiiii" (``let's go outttttt")\\
\indent A:``không đang ốm quá !!!!!!!! (`` no I am too sick !!!!!!"). \\
The intent of the utterance by person S that: he want to express more clearly his desire by using non-standard form ``iiiiiiiii" instead of the standard ``i". If the non-standard form was normalized to the standard form, in this case, the intent conveyed by the utterance would be ambiguous; ``iiiiiii" could suggest that person S is very excited to go out with person A. The non--standard word forms that contain additional pragmatic information presented in the non-standard form should be retained in the data pre--processing stage.
\item The message's short nature leading to the availability of very limited context information.   
\item In text chat dialogue, end of a turn is not always obvious. A turn often contains multiple messages. A message is often in a clause or utterance boundaries, but it is not always correct. Therefore, although the boundary of a message can be a useful feature to FS identification but sometimes a FS may contain multiple messages, and even may include only a part of one message and a part of the next message. This indistinct end of a turn also leads to the end of a misleading message. 
In sudden interruption cases, messages can become out of sync. Each participant tends to respond to a message earlier than the previous one, making the conversation also being out of order and the conversation seem inconsistent when read in sequence. This is a difficult problem for processing the dialog act segmentation.

\end{enumerate}

In short, unlike carefully authored news text, conversational text poses a number of new challenges, due to their short, context-dependent, noisy and dynamic nature. Tackling this challenge, ideally, requires changing related natural language processing tools to become suitable for texts from social media network or normalizing conversational texts to fit with existing tools. However, both of which are hard tasks. In the scope of this paper, we standardize the message data using our online chat dictionary to match popular abbreviations, acronyms, and slang with standard words in the pre-processing stage.
 
\subsection*{Online chat dictionary} Our online chat dictionary includes abbreviations, slang and the words that are written in teen style (teen code)  such as ``bj"- ``bây giờ" (``now"), ``ck" - ``chồng" (``husband"), ``4u" - ``cho bạn" (``for you"). The letters ``c", ``k", ``q" are usually replaced by ``k", ``ch” but often replaced with ``ck” ... Using online chat dictionary to standardize the message data, the noisiness of input data will be reduced. This make it more formal and help the models run better.

\subsection{Phone corpus}
Our Phone data set is build from scripted telephone speech of LDC2017S01 data (IARPA Babel Vietnamese Language Pack IARPA-babel107b-v0.7 \footnote{https://catalog.ldc.upenn.edu/LDC2017S01}). LDC2017S01 contains Vietnamese phone audios and transcripts. The Vietnamese conversations in these corpus contain different dialects that spoken in the North, North-Central, Central and Southern regions in Vietnam. We selected 22 conversations and segment its transcripts into the turn by manual. Then, the turns are annotated FS. The Phone data includes 1545 turns and 3500 FSs with an average of 70 turns and 160 FSs per conversation. The agreement scores of the phone data set is 0.84 Fleiss' kappa measure.\\
FS recognition for spoken texts, however, is more challenging than working with written documents due to some reasons as follows:
\begin{enumerate}
\item First, spoken text are commonly shorter and less grammatical, not comply with rigid syntactic constraints. Sentence elements like subject or object are often omitted. It is very context-dependent. Also, there are no punctuation marks in the texts. It, therefore, is non–trivial to segment and parse spoken sentences correctly.
\item Second, conversational speech contains a lot of self-correcting, hesitation, and stutter. This is one of the main reasons that causes nested FS. $fs_9$ and $fs_{13}$ within turn $t_4$ in Table \ref{Tab:phone} are the instances. 
\item Third, the output text of Automatic Speech Recognition are all in lowercase and bearing a small percentage of errors. 
\end{enumerate}
These challenges make it extremely difficult to recognize FS in particular and in understanding spoken language in general. 
\subsection*{Vietnamese local dialect dictionary}
The LDC2017S01 data is built from spoken conversations in the North, North-Central, Central and Southern Vietnamese dialect. Because of the nature of Vietnamese dialects, a lot of words in local dialects can be changed to standard dialect (the North Vietnamese dialect) without affecting the meaning of the utterances in which they belongs. For instances, ``Răng rứa" means ``sao thế" (what up); ``Mi đi mô" means ``Mày đi đâu" (where are you going?). Therefore we created a dictionary to match these words with standardized words. By doing so, the data sets become more uniform. This makes it easier to handle and help the models to run better. Our dictionary is not only useful in this study but also can be very helpful in all other studies that involve Vietnamese human--human, and human--machine conversation. 

\section{DA segmetation with ME, CRF and BI-LSTM-CRF} 
\label{sec:method}
The number of discontinuous or nested functional segments account for a very small percent in both data sets (0.5\% in the Message corpus, 0.9\% in the Phone corpus). Hence there are not enough discontinuous or nested functional segments so the models can learn to identify them. For that reason, this paper only focuses on identifying continuous and un-nested functional segments (which make up more than 99\% of both data sets). In future studies, we intend to increase the size of our data sets, the number of discontinuous or nested functional segments and study methods to identify these functional segments. In this paper, we cast the segmentation problem as a sequential tagging task: the first word of a FS is marked with B\_fs (Begin of a FS), the token that is inside of a FS is marked with I\_FS (Inside of a FS). The problem of FS identification in a sentence is modeled as the problem of labeling syllables in that sentence with two above labels. \\
Let $t= \left\{t_1, t_2,... t_n\right\}$ be turns and $y = \left\{B, I \right\}$ be per-token output tags. We predict the most likely y, given a conditional model $P(y|t)$.

\subsection{Maximum Entropy}
The ME (Maxent) model defines conditional distribution of class (y) given an observation vector t as the exponential form in Formula (\ref{1}) \cite{Berger:1996}: 
\begin{equation}\label{1}
{P}(y/t)=\frac{1}{Z(t)}exp\left(\sum_1^K\theta_{k}(t,y)\right)
\end{equation}
where $\theta_{k}$ is a weight parameter to be estimated for the corresponding feature function $f_k(t, y)$, and $Z(t)$ is a normalizing factor over all classes to ensure a proper probability. $K$ is the total number of feature functions. 
We decided to use ME for evaluation and comparison because it is commented that it is suitable for sparse data like natural language, encode various rich and overlapping features at different levels of granularity \cite{Nigam:1999}.

\subsection{Conditional Random Fields}
The CRFs model defines also the conditional distribution of the class (y) given an observation vector t as the Formular (\ref{1}) \cite{Lafferty:2001}. In which $\theta_{k}$ is a weight parameter to be estimated for the corresponding feature function $f_k(t, y)$, and $Z(t)$ is a normalizing factor over all classes to ensure a proper probability. And $K$ is the total number of feature functions. It is essentially a ME model over the entire sequence. 
It is unlike the Maxent above since it models the sequence information, because the Maxent model decides for each state independently with the other states. For example, a transcription utterance together with class tags used for the CRF word detection model in Dialog act segmentation as follows:

\renewcommand{\figurename}{Figure}
\begin{figure}[!ht]
\centering
\includegraphics[width=1.8in]{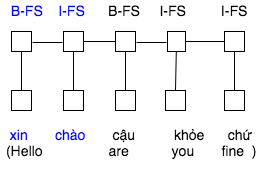}
\caption{A CRF model for identifying FS.}
\label{fig:CRF}
\end{figure}
Training ME and CRF are commonly performed by maximizing the likelihood function with respect to the training data using advanced convex optimization techniques like L--BFGS \cite{Liu:1989}.
\subsection{Deep learning--based models with Bi--LSTM--CRF}
Bi--LSTM--CRF network is formed by combining a bidirectional LSTM network and a CRF network \cite{Huang:2015}. Therefore Bi--LSTM--CRF can efficiently use past and future input features via a Bi--LSTM layer and sentence level tag information via a CRF layer. A CRF layer is represented by lines which connect consecutive output layers. A CRF layer has a state transition matrix as parameters. The following are examples of a text in the Bi--LSTM--CRF model:
\renewcommand{\figurename}{Figure}
\begin{figure}[!ht]
\centering
\includegraphics[width=2.9in]{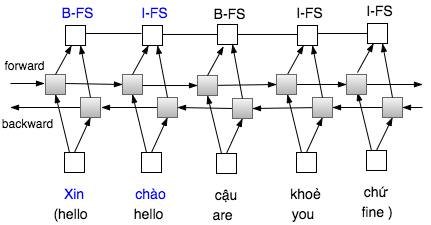}
\caption{A BI-LSTM-CRF model for identifying FS.}
\label{fig:LSTM}
\end{figure}
BI--LSTM--CRF has emerged as a standard method for obtaining per-token vector representations serving as input to various token labeling tasks. We expect that dialog act segmentation in Vietnamese using BI--LSTM--CRFs model will also similar to highly accurate results.  
\section{Evaluation}
\label{sec:evaluation}
The simple lexical feature, n--gram (unigram, bigram and trigram), is used for the ME and CRF models. We do experiments on two different conversational data sets (Message data set and Phone data set) after normalizing these data sets using local dialect dictionary and online chat dictionary. 

Training ME and CRF are commonly performed by maximizing the likelihood function with respect to the training data using quasi-Newton methods like L--BFGS \cite{Liu:1989}. Thus, in the experiments with ME and CRF, we use L-BFGS method. For CRF models, we use second-order Markov dependency. On experiment with CRF, we use tools: FlexCRFs - a C/C++ implementation of CRFs \footnote{http://flexcrfs.sourceforge.net/}. On experiment with Bi--LSTM--CRF, our setup is based on study of Lample et al.  \footnote{https://github.com/glample/tagger} \cite{Lample:2016} . 

For evaluating each experiments, we randomly divide each corpus into five parts to do 5-fold cross-validation test. In each fold we take one partition for testing and 4 partitions for training. The summary of the experiment results on Message data set is shown in Table \ref{Tab:result_Message}, the experiment results on Phone data set is shown in Table \ref{Tab:result_Phone}. \\
The results of label-based performance evaluation are significantly higher than the results of label-based performance evaluation and chunk-based performance evaluation. The evaluation measures for this task are precision and  recall based on labels:
\begin{center}
$precision = \frac{number\ of\ correctly\ predicted\ label\ by\  the \ model}{number\ of\ label\ predicted\ by\ the\ model}$;\\
\vspace{0.3cm}
$recall = \frac{number\ of\ correctly\ predicted\ label\ by\  the \ model}{number\ of\ actual\ label \ annotated\ by\ humans}$;\\
\end{center}
$Average_{macro}$ is the average of the precision and recall of the model on different classes. $Average_{micro}$ is sum up the individual true positives, false positives, and false negatives of the model for different classes. 

The precision and recall based on chunks is as follows: 
\begin{center}
$precision = \frac{number\ of\ correctly\ predicted\ FS\ by\  the \ model}{number\ of\ FS\ predicted\ by\ the\ model}$;\\
\vspace{0.3cm}
$recall = \frac{number\ of\ correctly\ predicted\ FS\ by\  the \ model}{number\ of\ actual\ FS \ annotated\ by\ humans}$; \\
\vspace{0.3cm}
\end{center}
$F_1$-- score in the both of evaluations is calculated as follows:
\begin{center}
$F_1 = \frac {2 * (precision * recall)}{(precision + recall)}$;\\
\end{center}
BI--LSTM--CRF models achieved the highest performance (average F1 of 90.42\% with Messages dataset, 73.26\% with Phone dataset). This was an indication that it is robust and less affected by the removal of engineering features.
\renewcommand\tablename{Table}
\begin{table}[ht]
\centering
\caption{Performance comparison among ME, CRF and BI--LSTM--CRF models on Message dataset.}
\label{Tab:result_Message}
\begin{tabular}{l|l|l|l|l}
\hline

\rowcolor[HTML]{EFEFEF} 
\textbf{Model}                & \textbf{Lable}                & \textbf{Precision}            & \textbf{Recall}               & \textbf{F1-score}                                             \\ \hline
                              & B-fs                          & 77.36                         & 77.74                         & 77.54                                                         \\ \cline{2-5} 
                              & I-fs                          & 94.67                         & 94.56                         & 94.61                                                         \\ \cline{2-5} 
                              & $Average_{macro}$                         & 86.01                         & 86.15                         & \textbf{86.08}                                                \\ \cline{2-5} 
                              & $Average_{micro}$                         & 91.31                         & 91.31                         & \textbf{91.31}                                                \\ \cline{2-5} 
\multirow{-5}{*}{ME}          & \cellcolor[HTML]{EFEFEF}\textbf{Chunk} & \cellcolor[HTML]{EFEFEF}57.38 & \cellcolor[HTML]{EFEFEF}57.33 & \cellcolor[HTML]{EFEFEF}\textbf{57.34}                        \\ \hline
                              & B-fs                          & 100                           & 80.03                         & 88.9                                                          \\ \cline{2-5} 
                              & I-fs                          & 95.46                         & 100                           & 97.68                                                         \\ \cline{2-5} 
                              & $Average_{macro}$                           & 97.73                         & 90.01                         & \textbf{93.71}                                                \\ \cline{2-5} 
                              & $Average_{micro}$                         & 96.16                         & 96.16                         & \textbf{96.16}                                                \\ \cline{2-5} 
\multirow{-5}{*}{CRF}         & \cellcolor[HTML]{EFEFEF}\textbf{Chunk} & \cellcolor[HTML]{EFEFEF}83.8  & \cellcolor[HTML]{EFEFEF}67.08 & \cellcolor[HTML]{EFEFEF}\textbf{74.51}                        \\ \hline
                              & B-fs                          & 97.11                         & 95.24                         & 96.17                                                         \\ \cline{2-5} 
                              & I-fs                          & 98.87                         & 99.32                         & 99.1                                                          \\ \cline{2-5} 
                              & $Average_{macro}$                          & 97.99                         & 97.28                         & \textbf{97.64}                                                \\ \cline{2-5} 
                              & $Average_{micro}$                         & 98.54                         & 98.54                         & \textbf{98.54}                                                \\ \cline{2-5} 
\multirow{-5}{*}{BI-LSTM-CRF} & \cellcolor[HTML]{EFEFEF}\textbf{Chunk} & \cellcolor[HTML]{EFEFEF}91.3  & \cellcolor[HTML]{EFEFEF}89.56 & \cellcolor[HTML]{EFEFEF}{\color[HTML]{3531FF} \textbf{90.42}} \\ \hline
\end{tabular}
\end{table}

\renewcommand\tablename{Table}
\begin{table}[ht]
\centering
\caption{Performance comparison among ME, CRF and BI--LSTM--CRF models on Phone dataset.}
\label{Tab:result_Phone}
\begin{tabular}{l|l|l|l|l}
\hline

\rowcolor[HTML]{EFEFEF} 
\textbf{Model}                & \textbf{Lable}                & \textbf{Precision}            & \textbf{Recall}               & \textbf{F1-score}                                             \\ \hline
                              & B-fs                          & 83.51                         & 75.89                         & 79.52                                                         \\ \cline{2-5} 
                              & I-fs                          & 93.9                          & 96.12                         & 95                                                            \\ \cline{2-5} 
                              & $Average_{macro}$                   & 88.7                          & 86.01                         & \textbf{87.34}                                                \\ \cline{2-5} 
                              & $Average_{micro}$                  & 91.96                         & 91.96                         & \textbf{91.96}                                                \\ \cline{2-5} 
\multirow{-5}{*}{ME}          & \cellcolor[HTML]{EFEFEF}\textbf{Chunk} & \cellcolor[HTML]{EFEFEF}61.88 & \cellcolor[HTML]{EFEFEF}56.23 & \cellcolor[HTML]{EFEFEF}\textbf{58.92}                        \\ \hline
                              & B-fs                          & 95.22                         & 71.24                         & 81.43                                                         \\ \cline{2-5} 
                              & I-fs                          & 93.09                         & 99.06                         & 95.98                                                         \\ \cline{2-5} 
                              & $Average_{macro}$                   & 94.15                         & 85.15                         & \textbf{89.42}                                                \\ \cline{2-5} 
                              & $Average_{micro}$                  & 93.4                          & 93.34                         & \textbf{93.37}                                                \\ \cline{2-5} 
\multirow{-5}{*}{CRF}         & \cellcolor[HTML]{EFEFEF}\textbf{Chunk} & \cellcolor[HTML]{EFEFEF}66.82 & \cellcolor[HTML]{EFEFEF}50.18 & \cellcolor[HTML]{EFEFEF}\textbf{57.27}                        \\ \hline
                              & B-fs                          & 94.38                         & 84.6                          & 89.22                                                         \\ \cline{2-5} 
                              & I-fs                          & 96.02                         & 98.64                         & 97.31                                                         \\ \cline{2-5} 
                              & $Average_{macro}$                   & 95.2                          & 91.62                         & \textbf{93.38}                                                \\ \cline{2-5} 
                              & $Average_{micro}$                  & 95.7                          & 95.7                          & \textbf{95.7}                                                 \\ \cline{2-5} 
\multirow{-5}{*}{BI-LSTM-CRF} & \cellcolor[HTML]{EFEFEF}\textbf{Chunk} & \cellcolor[HTML]{EFEFEF}77.48 & \cellcolor[HTML]{EFEFEF}69.47 & \cellcolor[HTML]{EFEFEF}{\color[HTML]{3166FF} \textbf{73.26}} \\ \hline
\end{tabular}
\end{table}
Performance results with Messages data (manual texts) are higher than results achieved with Phones data (Automatic Speech Recognition transcripts) because turns in Messages data set are often shorter and less ambiguous for dialog act segmentation than turns in Phone data set. Turns in Phone data set also includes hesitance, repeat, and overlap. These make discontinuous segments, either within a turn or spread over several turns as we have already discussed. A greater challenge is posed by those cases where different functional segments overlapped.

Another observation from the results is that Bi-LSTM-CRFs, the deep learning approach, performs significantly better than both CRF and ME, the machine learning approaches, by every measure.  Because deep learning has never been used for dialog act segmentation before, this result opens up a very promising new direction for future studies to approach dialog act segmentation and dialog act in general. Between the machine learning approaches, CRF performs better than ME overall. This can be explained by looking at how CRF and ME works. ME is locally re-normalized and suffers from the label bias problem, while CRFs are globally re-normalized. This label bias problem can happen a lot, especially with very context-dependent data sets like Message corpus and Phone corpus.


\section{Conclusions}
\label{Sec:concl}

We have presented a thorough investigation on Vietnamese FS identification using machine learning approach and deep learning approach. We built two annotated corpora for evaluation and two dictionaries that make the data sets more uniform and help the models run better. Two machine learning techniques and a deep learning technique are applied and compared on the task of automatic dialog act segmentation. Deep learning technique is also applied for the first time to dialog act segmentation. We also draw some useful conclusions observed from the experimental results that can be very helpful for future studies. 

These encouraging results show that the task of identifying functional segment is promising to continue to the next dialogue act identification steps and towards understanding intentions in the users’ utterances for Vietnamese. For future work, we intend to extend the studies into two directions. First, we plan to increase the size of our data set to get sufficient amount of instances in different types of functional segment and study deeper methods to solve nested FS identification. Second, we intend to use features included in the data sets as dialogue history, prosody to improve automatic FSs recognition and dialogue processing.




%



\section*{Acknowledgment}
This work was supported by the project QG.15.29 from Vietnam National University, Hanoi (VNU).




%

\renewcommand\refname{Reference}

%
%

\end{document}